\pdfoutput=1
\documentclass[11pt,a4paper]{article}
\usepackage[hyperref]{acl2017}
\usepackage{times}
\usepackage{latexsym}
\newcommand{\utt}[1]{``\textsl{#1}''}
\usepackage{color}

\usepackage{url}
\usepackage{multirow}
\usepackage{graphicx}
\usepackage{amsbsy}
\usepackage{arydshln}
\usepackage{linguex}

\aclfinalcopy % Uncomment this line for the final submission
\title{Not All Dialogues are Created Equal: \\ Instance Weighting for Neural Conversational Models}

\author{Pierre Lison \\
  Norwegian Computing Center \\
  Oslo, Norway \\
  {\tt plison@nr.no} 
  \And
  Serge Bibauw\thanks{\ \ Also affiliated with Universidad Central del Ecuador (Quito, Ecuador).} \\
  KU Leuven, imec \\
  Universit\'e catholique de Louvain \\
  {\tt serge.bibauw@kuleuven.be} \\ }

\date{}
\begin{document}
\maketitle
\begin{abstract}
  Neural conversational models require substantial amounts of dialogue data to estimate their parameters and are therefore usually learned on large corpora such as chat forums, Twitter discussions or movie subtitles. These corpora are, however, often challenging to work with, notably due to their frequent lack of turn segmentation and the presence of multiple references external to the dialogue itself. 
  This paper shows that these challenges can be mitigated by adding a \textit{weighting model} into the neural architecture. The weighting model, which is itself estimated from dialogue data, associates each training example to a numerical weight that reflects its intrinsic quality for dialogue modelling. At training time, these sample weights are included into the empirical loss to be minimised. Evaluation results on retrieval-based models trained on movie and TV subtitles demonstrate that the inclusion of such a weighting model improves the model performance on unsupervised metrics. 
\end{abstract}

\section{Introduction}

The development of conversational agents (such as mobile assistants, chatbots or interactive robots) is increasingly based on data-driven methods aiming to infer conversational patterns from dialogue data. One major trend in the last recent years is the emergence of neural conversation models \cite{vinyals2015neural,DBLP:conf/naacl/SordoniGABJMNGD15,shang-lu-li:2015:ACL-IJCNLP,Serban:2016:BED:3016387.3016435,lowe-dialog-ubuntu-2017,li2017adversarial}. These neural models can be directly estimated from raw (non-annotated) dialogue corpora, allowing them to be deployed with a limited amount of domain-specific knowledge and feature engineering. 

Due to their large parameter space, the estimation of neural conversation models requires considerable amounts of dialogue data. They are therefore often trained on conversations collected from various online resources, such as Twitter discussions \cite{Ritter:2010:UMT:1857999.1858019} online chat logs \cite{lowe-dialog-ubuntu-2017}, movie scripts \cite{Danescu-Niculescu-Mizil+Lee:11a} and movie and TV subtitles \cite{opensubtitles2016}.  

Although these corpora are undeniably useful, they also face some limitations from a dialogue modelling perspective. First of all, several dialogue corpora, most notably those extracted from subtitles, do not include any explicit turn segmentation or speaker identification \cite{serban2015text,slt2016}. In other words, we do not know whether two consecutive sentences are part of the same dialogue turn or were uttered by different speakers. The neural conversation model may therefore inadvertently learn responses that remain within the same dialogue turn instead of starting a new turn. 

Furthermore,  these dialogues contain multiple references to named entities (in particular, person names such as fictional characters) that are specific to the dialogue in question.  These named entities should ideally not be part of the conversation model, since they often draw on an external context that is absent from the inputs provided to the conversation model. For instance, the mention of character names in a movie is associated with a visual context (for instance, the characters  appearing in a given scene) that is not captured in the training data. 
Finally, a substantial portion of the utterances observed in these corpora is made of neutral, commonplace responses (\utt{Perhaps}, \utt{I don't know}, \utt{Err}, ...) that can be used in most conversational situations but fall short of creating  meaningful and engaging conversations with human users \cite{li-EtAl:2016:N16-11}.

The present paper addresses these limitations by adding a \textit{weighting model} to the neural architecture. The purpose of this model is to associate each $\langle \textit{context},\textit{response} \rangle$ example pair to a numerical {\it weight} that reflects the intrinsic ``quality'' of each example. The instance weights are then included in the empirical loss to minimise when learning the parameters of the neural conversation model. The weights are themselves computed via a neural model learned from dialogue data. Experimental results demonstrate that the use of instance weights improves the performance of neural conversation models on unsupervised metrics. Human evaluation results are, however, inconclusive. 

The rest of this paper is as follows. The next section presents a brief overview of existing work on neural conversation models. Section \ref{sec:approach} provides a description of the instance weighting approach. Section \ref{sec:evaluation} details the experimental validation of the proposed model, using both unsupervised metrics and a human evaluation of the selected responses. Finally, Section \ref{sec:discussion} discusses the advantages and limitations of the approach, and Section \ref{sec:conclusion} concludes this paper.

\section{Related Work}
\label{sec:background}

Neural conversation models are a family of neural architectures (generally based on deep convolutional or recurrent networks) used to represent mappings between dialogue contexts (or queries) and possible responses. Compared to previous statistical approaches to dialogue modelling based on Markov processes \cite{817450,RieserLemon11,6407655}, one benefit of these neural models is their ability to be estimated from raw dialogue corpora,  without having to rely on additional annotation layers for intermediate representations such as state variables or dialogue acts. Rather, neural conversation models automatically {\it derive} latent representations of the dialogue state based on the observed utterances. 

Neural conversation models can be divided into two main categories, {\it retrieval models} and {\it generative models}. Retrieval models are used to select the most relevant response for a given context amongst a (possibly large) set of predefined responses, such as the set of utterances extracted from a corpus \cite{DBLP:conf/sigdial/LowePSP15,DBLP:journals/corr/PrakashBA16}. Generative models, on the other hand, rely on sequence-to-sequence models \cite{DBLP:conf/naacl/SordoniGABJMNGD15} to generate new, possibly unseen responses given the provided context. These models are built by linking together two recurrent architectures: one encoder which maps the sequence of input tokens in the context utterance(s) to a fixed-sized vector, and one decoder that generates the response token by token given the context vector \cite{vinyals2015neural,DBLP:conf/naacl/SordoniGABJMNGD15}. Recent papers have shown that the performance of these generative models can be improved by incorporating attentional mechanisms \cite{DBLP:journals/corr/YaoPZW16} and accounting for the structure of conversations through hierarchical networks \cite{Serban:2016:BED:3016387.3016435}. Neural conversation models can also be learned using adversarial learning \cite{li2017adversarial}. In this setting, two neural models are jointly learned: a generative model producing the response, and a discriminator optimised to distinguish between human-generated responses and machine-generated ones. The discriminator outputs are then used to bias the generative model towards producing more human-like responses.

The linguistic coherence and diversity of the models can be enhanced by including speaker-addressee information \cite{li-EtAl:2016:P16-13} and by expressing the objective function in terms of Maximum Mutual Information to enhance the diversity of the generated responses \cite{li-EtAl:2016:N16-11}. As demonstrated by \cite{DBLP:journals/corr/GhazvininejadBC17}, neural conversation models can also be combined with external knowledge sources in the form of factual information or entity-grounded opinions, which is an important requirement for developing task-oriented dialogue systems that must ground their action in an external context. 

Dialogue is a sequential decision-making process where the conversational actions of each participant influence not only the current turn but the long-term evolution of the dialogue \cite{817450}. To incorporate the prediction of future outcomes in the generation process, several papers have explored the use of reinforcement learning techniques, using deep neural networks to model the expected future reward \cite{DBLP:journals/corr/LiMRGGJ16,Cuayahuitl2017}. In particular, the Hybrid Code Networks model of \cite{DBLP:journals/corr/WilliamsAZ17} demonstrate how a mixture of supervised learning, reinforcement learning and domain-specific knowledge can be used to optimise dialogue strategies from limited amount of training data.

In contrast with the approaches outlined above, this paper does not present a new neural architecture for conversational models. Rather, it investigates how the performance of existing models can be improved ``upstream'', by adapting how these models can be trained on large, noisy corpora with varying levels of quality. It should be noted that, although the experiments presented in Section \ref{sec:evaluation} focus on a limited range of neural models, the approach presented in this paper is designed to be model-independent and can be applied as a preprocessing step to any data-driven model of dialogue. 

\section{Approach}
\label{sec:approach}

\begin{figure*}[t]
\begin{center}
\includegraphics[scale=0.42]{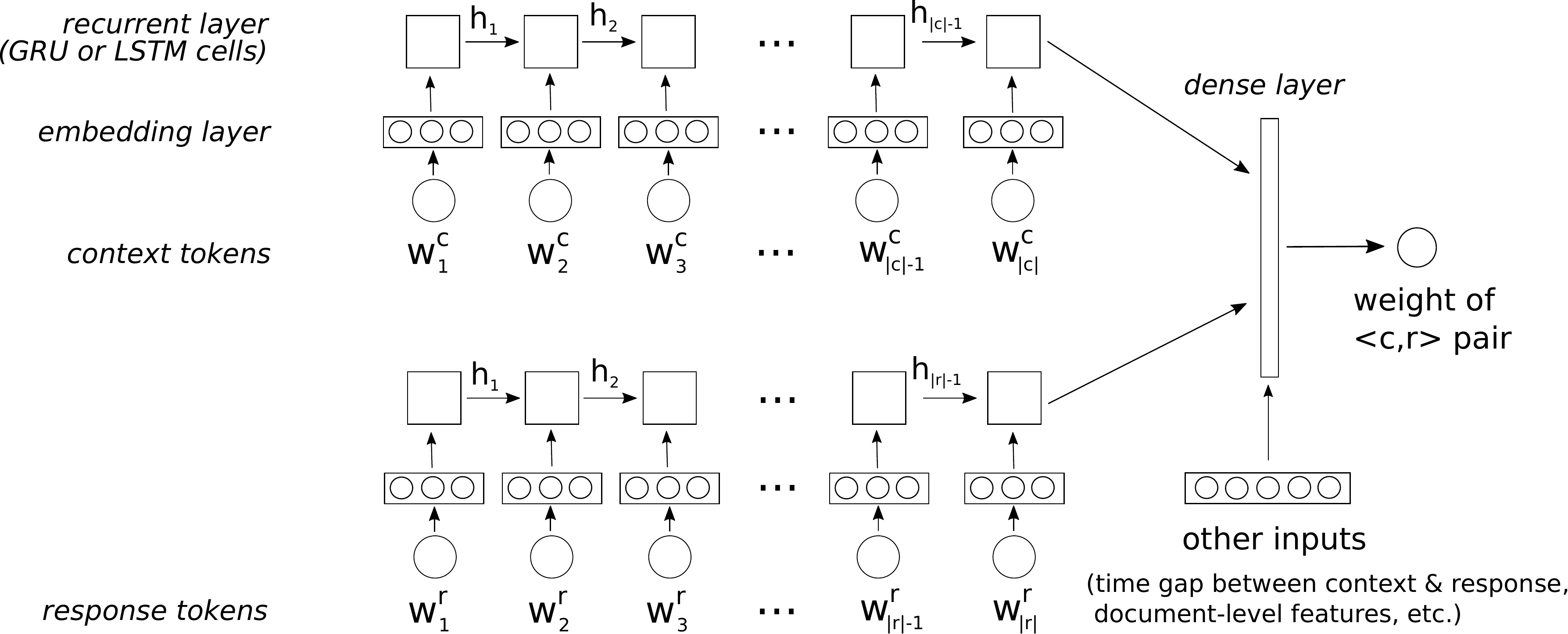}
\end{center}
\caption{Neural weighting model, taking as input the $\langle \textrm{context}, \textrm{response} \rangle$ pairs, possibly along additional features (such as timing information for subtitles), and returning an associated weight value.}
\label{fig:weightmodel}
\end{figure*}

As mentioned in the introduction, the interactions extracted from large dialogue corpora do not all have the same intrinsic quality, due for instance to the frequent lack of turn segmentation or the presence of external, unresolvable references to person names. In other words, there is a discrepancy between the actual $\langle \textrm{context}, \textrm{response} \rangle$ pairs found in these corpora and the conversational patterns that should be accounted for in the neural model. 

One way to address this discrepancy is by framing the problem as one of {\it domain adaptation}, the source domain being the original dialogue corpus and the target domain representing the dialogues we want our model to produce. The target domain is in this case not necessarily another dialogue domain, but simply reflects the fact that the distribution of responses in the raw corpus does not necessarily reflect the distribution of responses we ultimately wish to encode in the conversational model.  

A popular strategy for domain adaptation in natural language processing, which has notably been used in POS-tagging, sentiment analysis, spam filtering and machine translation \cite{Bickel:2007:DLD:1273496.1273507,citeulike:2638020,Foster:2010:DIW:1870658.1870702,Xia:2013:ISI:2540128.2540441}, is to assign a higher weight to training instances whose properties are similar to the target domain. We present below such an instance weighting approach tailored for neural conversational models.

\subsection{Weighting model}
\label{sec:weighting_model}

The quality of a particular $\langle \textrm{context}, \textrm{response} \rangle$ pair is difficult to determine using handcrafted rules -- for instance, the probability of a turn boundary may depend on multiple factors such as the presence of turn-yielding cues or the time gap between the utterances \cite{slt2016}. To overcome these limitations, we adopt a data-driven approach and automatically learn a weighting model from examples of ``high-quality'' responses. What constitutes a high-quality response depends in practice on the specific criteria we wish to uphold in the conversation model -- for instance, favouring responses that are likely to form a new dialogue turn (rather than a continuation of the current turn),  avoiding the use of dull, commonplace responses, or disfavouring the selection of responses that contain unresolved references to person names.

 The weighting model can be expressed as a neural model which associates each $\langle \textrm{context}, \textrm{response} \rangle$ example pair to a numerical weight. The architecture of this neural network is depicted in Figure \ref{fig:weightmodel}. It is composed of two recurrent sub-networks with shared weights, one for the context and one for the response. Each sub-network takes a sequence of tokens as input and pass them through an embedding layer and a recurrent layer with LSTM or GRU cells. The fixed-size vectors for the context and response are then fed to a regular densely-connected layer, and finally to the final weight value through a sigmoid activation function. Additional features can also be included  whenever available -- for instance, timing information for movie and TV subtitles (such as the duration gap  between the context and its response, in milliseconds), or document-level features such as the dialogue genre or the total duration of the dialogue. 

To estimate its parameters, the neural model is provided with positive examples of ``high-quality'' responses along with negative examples sampled at random from the corpus. Based on this training data, the network learns to assign higher weights to the $\langle \textrm{context}, \textrm{response} \rangle$ pairs whose output vectors (combined with the additional inputs) are close from the high-quality examples, and a lower weight for those further away. In practice, the selection of high-quality example pairs from a given corpus can be performed through a combination of simple heuristics, as detailed in Section \ref{sec:models}.

\subsection{Instance weighting}
\label{sec:instance_weighting}

Once the weighting model is estimated, the next step is to run it on the entire dialogue corpus to compute the expected weight of each $\langle \textrm{context}, \textrm{response} \rangle$ pair. These sample weights are then included in the empirical loss that is being minimised during training. Formally, assuming a set of context-response pairs $\{(c_1,r_1), (c_2,r_2), ... (c_n,r_n)\}$ with associated weights $\{w_1,...w_n\}$, the estimation of the model parameters $\boldsymbol\theta$ is expressed as a minimisation problem.  For retrieval models, this minimisation is expressed as:
\begin{equation}
\boldsymbol\theta^* = min_{\boldsymbol\theta} \sum_1^n w_i \ L(y_i, f(c_i,r_i ; \boldsymbol\theta )  )
\end{equation}

where $L$ is a loss function (for instance, the cross-entropy loss), and $y_i$ is set to either 1 if $r_i$ is the response to $c_i$, and 0 otherwise (when $r_i$ is a negative example). For generative models, the minimisation is similarly expressed as:
\begin{equation}
\boldsymbol\theta^* = min_{\boldsymbol\theta} \sum_1^n w_i  \ L(r_i, f(c_i ; \boldsymbol\theta ) )
\end{equation}

In both cases, the loss computed from each example pair is multiplied by the weight value determined by the weight model. Examples associated with a larger weight $w_i$ will therefore have a larger influence on the gradient update steps.

\section{Evaluation}
\label{sec:evaluation}

The approach is evaluated on the basis of retrieval-based neural models trained on English-language subtitles from \cite{opensubtitles2016}. 
Three alternative models are evaluated:
\begin{enumerate}
    \item A traditional TF-IDF model, 
    \item A Dual Encoder model trained directly on the corpus examples, 
    \item A Dual Encoder model combined with the weighting model from Section \ref{sec:weighting_model}. 
\end{enumerate}

%Movie and TV subtitles are particularly interesting for neural conversational models, due to their size and linguistic diversity\footnote{The latest release of the OpenSubtitles collection of corpora contains over 3 million subtitles covering 60 languages and a large variety of genres and registers.}.

    % The reliance on retrieval-based neural models for this evaluation is motivated by the fact that they are relatively easy to train, yet can encode complex dialogue patterns.

\subsection{Models}

\label{sec:models}

\begin{figure*}[t!]
\begin{center}
\includegraphics[scale=0.42]{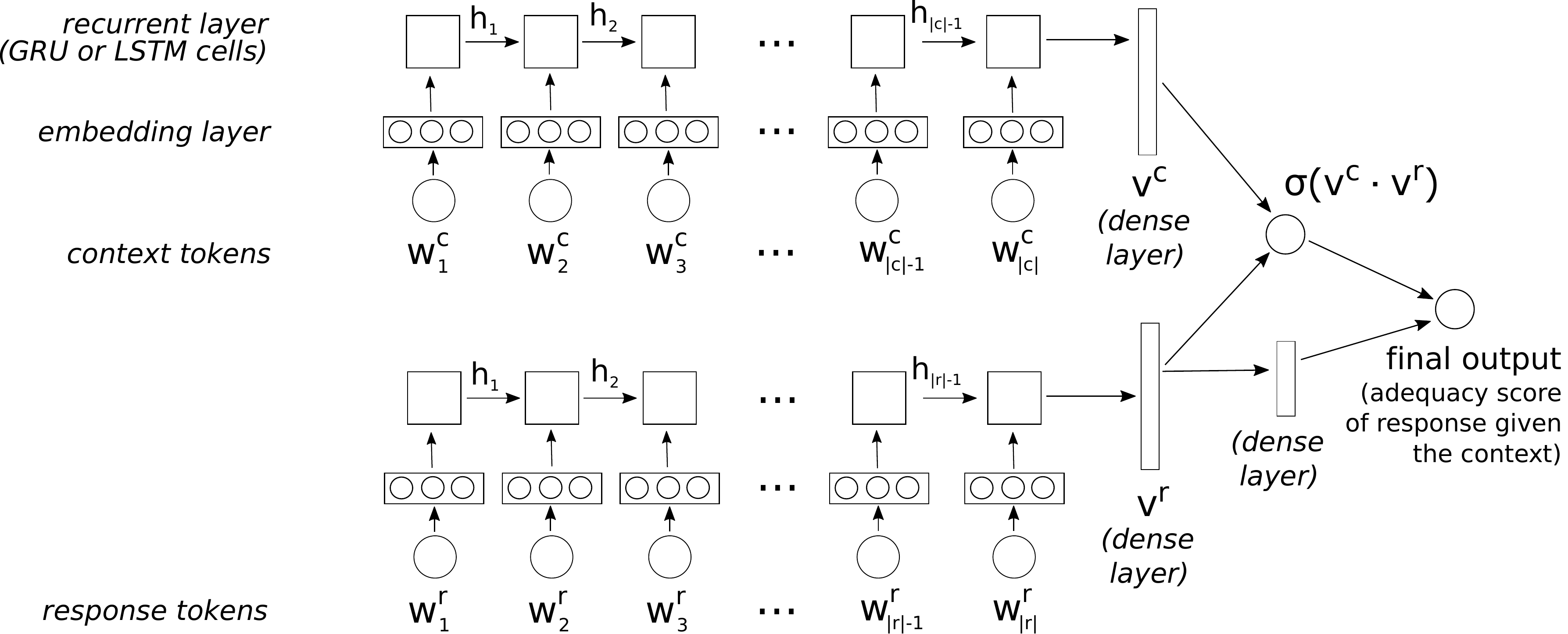}
\end{center}
\caption{Dual encoder model, taking as input the $\langle \textrm{context}, \textrm{response} \rangle$ pairs and returning a score expressing the adequacy of the response given the context. }
\label{fig:dualencoder}
\end{figure*}

\subsubsection*{TF-IDF model}

The TF-IDF (Term Frequency - Inverse Document Frequency) model computes the similarity between the context and its response using methods from information retrieval \cite{ramos2003using}. TF-IDF  measures the importance of a word in a ``document'' (in this case the context or response) relative to the whole corpus.  The model transforms the context and response (represented as bag-of-words) into TF-IDF-weighted vectors. These vectors are sparse vectors of a size equivalent to the vocabulary size, where each row corresponds, if the given word is present in the context or response, to its TF-IDF weight, and is 0 otherwise. The matching score between the context and its response is then determined as the cosine similarity between the two vectors:
\begin{equation}
\textit{similarity} = \frac{v^c \cdot v^ r}{\left\|v^c\right\|_2 \left\|v^r\right\|_2}
\end{equation}
where $v^c$ and $v^r$ respectively denote the TF-IDF-weighted vectors for the context and response.

\subsubsection*{Dual Encoder}

The Dual Encoder model \cite{lowe-dialog-ubuntu-2017} consists of two recurrent networks, one for the context and one for the response. The tokens are first passed through an embedding layer and then to a recurrent layer with LSTM or GRU cells. In the original formalisation of this model \cite{DBLP:conf/sigdial/LowePSP15}, the context vector is transformed through a dense layer of same dimension, representing the ``predicted'' response. The inner product of the predicted and actual responses is then calculated and normalised, yielding a similarity score. This model, however, only seeks to capture the semantic similarity between the two sequences, while the selection of the most adequate response in a given context may also need to account for other factors such as the grammaticality and coherence of the response. We therefore extend the Dual Encoder model in two ways. First, both the context and response vectors are transformed through a dense layer at the end of the recurrent layer (instead of just the context vector). Second, the final prediction is connected to both the inner product of the two vectors and to the response vector itself, as depicted in Figure \ref{fig:dualencoder}.

\subsubsection*{Dual Encoder with instance weighting}

Finally, the third model relies on the exact same Dual Encoder model as above, but applies the weighting model described in Section \ref{sec:weighting_model} prior to learning in order to assign weights to each training example. The weighting model is estimated on a subset of the movie and TV subtitles augmented with speaker information and filtered through heuristics to ensure a good cohesion between the context and its response. These heuristics are detailed in the next section. 

Although the architecture of the Dual Encoder is superficially similar to the weighting model of Figure \ref{fig:weightmodel}, the two models serve a different purpose: the weighting model returns the expected \textit{quality} of a training example, while the Dual Encoder returns a score expressing the \textit{adequacy} between the context and the response.

\subsection{Datasets}
\label{sec:datasets}

\subsubsection*{Training data for the conversation models}

The dataset used for training the three retrieval models is the English-language portion of the OpenSubtitles corpus of movie and TV subtitles \cite{opensubtitles2016}. The full dataset is composed of 105 445 subtitles and 95.5 million utterances, each utterance being associated with a start and end time (in milliseconds).  

\subsubsection*{Training data for the weighting model}

For training the weighting model, we extracted a small subset of the full corpus of subtitles corresponding to $\langle \textrm{context}, \textrm{response} \rangle$ pairs satisfying specific quality criteria. The first step was to align at the sentence level the subtitles with an online collection of movie and TV scripts (1 069 movies and 6 398 TV episodes), following the approach described in \cite{slt2016}.

This alignment enabled us to annotate the subtitles with speaker names and turn boundaries. Based on these subtitles, we then selected example pairs with two heuristics:
\begin{enumerate}
\item To ensure the response constitutes an actual reply from another speaker and not simply a continuation of the current turn, the subtitles were segmented into sub-dialogues. $\langle \textrm{context}, \textrm{response} \rangle$ pairs including a change of speaker from the context to the response were then extracted from these sub-dialogues. Since multi-party dialogues make it harder to determine who replies to whom, only sub-dialogues with two participants were considered in the subset. 
\item To ensure the response is intelligible given the context (without drawing on unresolved references to e.g. fictional person names), we also filtered out from the subset the dialogue turns including mentions of fictional character names and out-of-vocabulary words.
\end{enumerate}

A total of 95 624 $\langle \textrm{context}, \textrm{response} \rangle$ pairs can be extracted using these two heuristics. This corresponds to about 0.1 \% of the total number of examples for the OpenSubtitles corpus. These pairs are used as positive examples for the weighting model, along with negative pairs sampled at random from the corpus. 

\subsubsection*{Test data}

Two distinct corpora are used as test sets for the evaluation. The first corpus, whose genre is relatively close to the training set, is the Cornell Movie Dialog Corpus \cite{Danescu-Niculescu-Mizil+Lee:11a}, which is a collection of fictional conversations extracted from movie scripts (unrelated to the ones used for training the weighting model). The transcripts from this corpus are segmented into conversations. Each conversation is represented as a sequence of dialogue turns. As this paper concentrates on the selection of relevant responses in a given context, we limited the test pairs to the ones where the context ends with a question, which yields a total of 67 305 $\langle \textrm{context}, \textrm{response} \rangle$ pairs. 

The second test set comes from a slightly different conversational genre, namely theatre plays. The scripts of 62 English-language theatre plays were downloaded from public websites. We also limited the test pairs to the pairs where the context ends with a question, for a total of 3 427 pairs. 

\subsubsection{Experimental design}
\label{sec:unsupervised}

\subsubsection*{Preprocessing}

The utterances from all datasets were tokenised, lemmatised and POS-tagged using the spaCy NLP library\footnote{\textsf{\url{https://spacy.io/}}}. We also ran the named entity recogniser from the same library to extract named entities. Since the person names mentioned in movies and theatre plays typically refer to fictional characters, we replaced their occurrences by tags, one distinct tag per entity. For instance, the pair: \vspace{2mm}

\hspace{-5mm} \begin{tabular} {lp{54mm}}
 {\bf Dana}: & Frank, do you think you could give me a hand with these bags? \\ 
 {\bf Frank}: & I'm not a doorman, Miss Barrett.  I'm a building superintendent. \\ \vspace{-2mm}
 \end{tabular}

is simplified as: \vspace{2mm}

\hspace{-5mm}  \begin{tabular} {lp{54mm}}
 {\bf Dana}: & \begin{small}\textsf{\textless person1\textgreater}\end{small}, do you think you could give me a hand with these bags? \\ 
 {\bf Frank}: & I'm not a doorman, \begin{small}\textsf{\textless person2\textgreater}\end{small}.  I'm a building superintendent. \\ \vspace{0mm}
 \end{tabular}
 
Named entities of locations and numbers are also replaced by similar tags.  To account for the turn structure, turn boundaries were annotated with a \begin{small}\textsf{\textless newturn\textgreater}\end{small} tag. The vocabulary is capped to 25 000 words determined from their frequency in the training corpus. Tokens not covered in this vocabulary are replaced by  \begin{small}\textsf{\textless unknown\textgreater}\end{small}.  
 
\subsubsection*{Training details}

The dialogue contexts were limited to the last 10 utterances preceding the response and a maximum of 60 tokens.  The responses were defined as the next dialogue turn after the context, and limited to a maximum of 5 utterances and 30 tokens.

\begin{table*}[t!]
\begin{center}
\begin{tabular}{p{42mm}|rrr|rrr} \hline
Model name & \multicolumn{3}{c|}{Cornell Movie Dialogs} & \multicolumn{3}{c}{Theatre plays} \\
& $\textrm{R}_{10}@1$ &  $\textrm{R}_{10}@2$ &  $\textrm{R}_{10}@5$ &  $\textrm{R}_{10}@1$ &  $\textrm{R}_{10}@2$ &  $\textrm{R}_{10}@5$\\ \hline
TF-IDF & 0.33 & 0.44 & 0.67  & 0.33 & 0.44 & 0.53 \\
Dual Encoder & 0.44 & 0.62 & 0.83 & 0.52 & 0.67  & 0.75 \\
Dual Encoder + weighting & {\bf 0.47} & {\bf 0.63} & {\bf 0.85} & {\bf 0.56} & {\bf 0.70} & {\bf 0.80} \\ \hline
\end{tabular}
\end{center}
\caption{Performance of the 3 retrieval models on the two test sets, namely the Cornell Movie Dialogs Dataset and the smaller dataset of theatre plays, using the $\textrm{Recall}_{10}@i$ metric.}
\label{table:results}
\end{table*}

The embedding layers of the Dual Encoders were initialised with Skip-gram embeddings trained on the OpenSubtitles corpus. For the recurrent layers, we tested the use of both GRU and LSTM cells, along with their bidirectional equivalents  \cite{DBLP:journals/corr/ChungGCB14}, without noticeable differences in accuracy.  As GRU cells are faster to train than LSTM cells, we opted for the use of GRU-based recurrent layers. The dimensionality of the output vectors from the recurrent layers was 400. The neural networks are trained with a batch size of 256, binary cross-entropy as cost function and RMSProp as optimisation algorithm. To avoid overfitting issues, a dropout of 0.2 was applied at all layers of the neural model.  

Both the weighting model and the Dual Encoder models were training with a 1:1 ratio between positive examples (actual $\langle$ context, response $\rangle$ pairs) and negative examples with a response sampled at random from the training set.

\subsection{Results}

The three models (the TF-IDF model, the baseline Dual Encoder and the Dual Encoder combined with the weighting model) are evaluated using the Recall$_m@i$ metric, which is the most common metric for the evaluation of retrieval-based models. Let $\{ \langle c_i, r_i \rangle, 1 \leq i \leq n \}$ be the list of $m$ context-response pairs from the test set. For each context $c_i$,  we create a set of $m$ alternative responses, one response being the actual response $r_i$, and the $m\!-\!1$ other responses being sampled at random from the same corpus. The $m$ alternative responses are then ranked based on the output from the conversational model, and the Recall$_m@i$ measures how often the correct response appears in the top $i$ results of this ranked list.  The  Recall$_m@i$  metric is often used for the evaluation of retrieval models as several responses may be equally ``correct'' given a particular context.

The experimental results are shown in Table \ref{table:results}. As detailed in the table, the Dual Encoder model combined with the weighting model outperforms the Dual Encoder baseline on both test sets (the Cornell Movie Dialogs corpus and the smaller corpus of theatre plays). Our hypothesis is that the weighting model biases the responses selected by the conversation model towards more cohesive adjacency pairs between context and response\footnote{Contrary to the OpenSubtitles corpus which is made of subtitles with no turn segmentation, the Cornell Movie Dialogs corpus and the corpus of theatre plays are derived from scripts and are therefore segmented in dialogue turns.}. 

\begin{figure}[h]
\begin{center}
\includegraphics[scale=0.38]{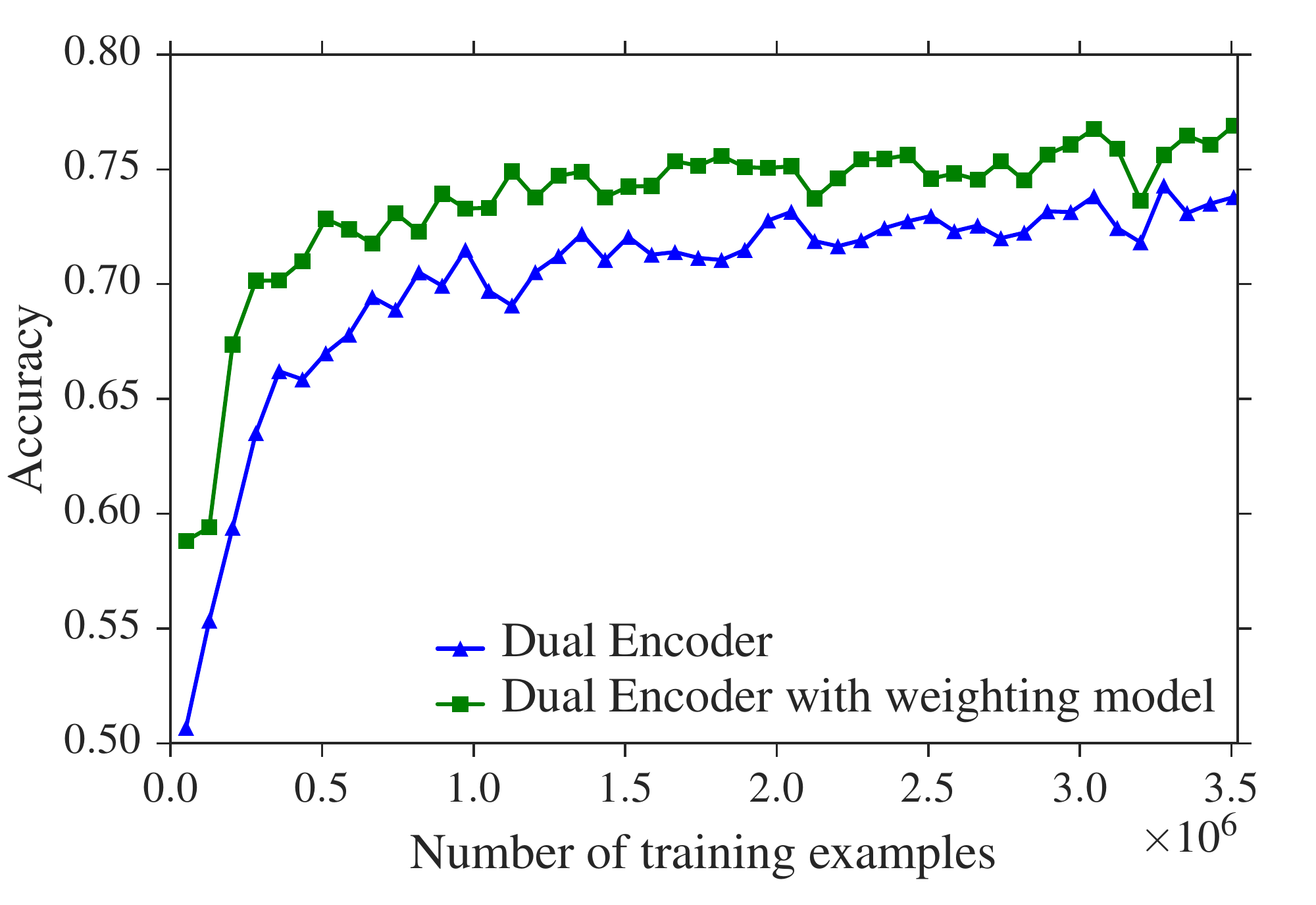} \vspace{-8mm}
\end{center}
\caption{Learning curve for the two Dual Encoder models, showing the evolution of their accuracy on the validation set as a function of the number of observed training examples.}
\label{fig:learningcurve}
\end{figure}

Figure \ref{fig:learningcurve} illustrates the learning curve for the two Dual Encoder models, where the accuracy is measured on a validation set composed of the high-quality example pairs described in the previous section along with randomly sampled alternative responses (using a 1:1 ratio of positive vs. negative examples). We can observe that the Dual Encoder with instance weights outperforms the baseline model on this validation set -- which is not \textit{per se} a surprising result, since the purpose of the weighting model is precisely to bias the conversation model to give more importance to these types of example pairs. 

\subsection{Human evaluation}

To further investigate the potential of this weighting strategy for neural conversational models, we conducted a human evaluation of the responses generated by the two neural models included in the evaluation. We collected human judgements on $\langle \textrm{context}, \textrm{response} \rangle$ pairs using a crowdsourcing platform. We extracted 115 random contexts from the Cornell Movie Dialogs corpus  and used four distinct strategies to generate dialogue responses: a random predictor (used to identify the lower bound), the two Dual Encoder models (both without and with instance weights), and expert responses (used to identify the upper bound). The expert responses were manually authored by two human annotators. The resulting 460 $\langle \textrm{context}, \textrm{response} \rangle$ pairs were evaluated by 8 distinct human judges each (920 ratings per model). The human judges were asked to rate the consistency between context and response on a 5-points scale, from \textit{Inconsistent} to \textit{Consistent}. In total, 118 individuals participated in the crowdsourced evaluation.

\begin{figure}[htbp]
\begin{center}
\hspace{-4mm}
\includegraphics[scale=0.7]{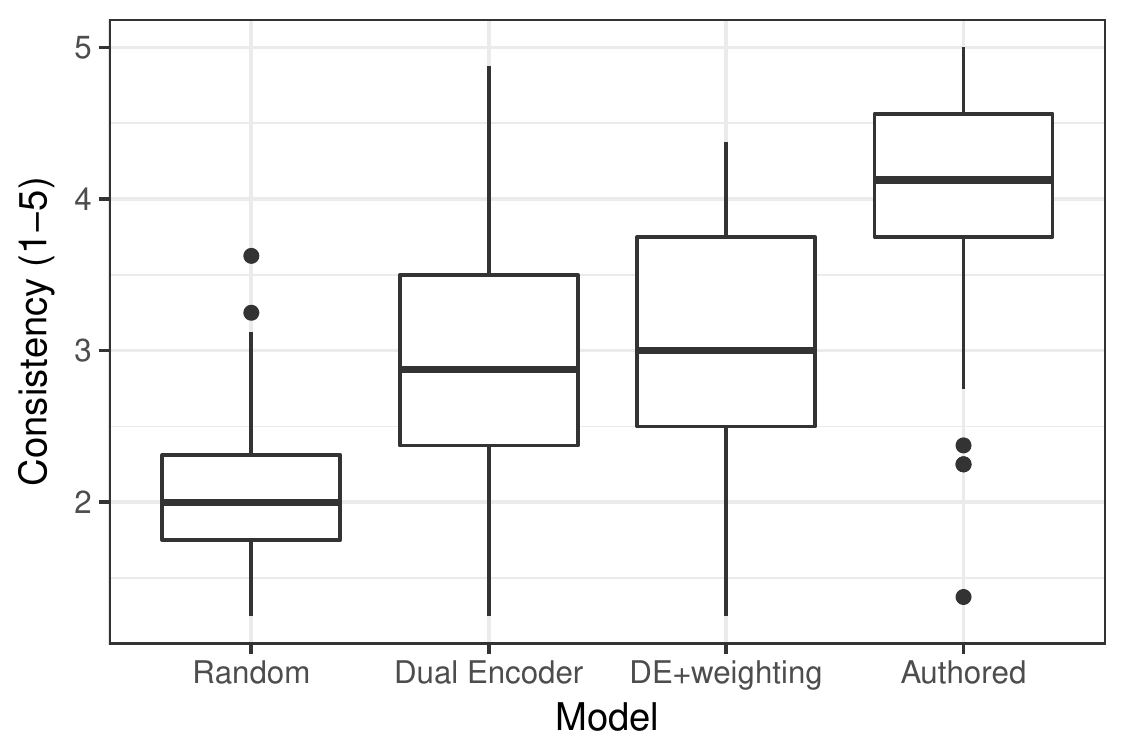} \vspace{-8mm} 
\end{center}
\caption{Distribution of human ratings of the responses generated by the four models tested.}
\label{fig:HumanEvalResults}
\end{figure}

The results of this human evaluation are presented in Figure \ref{fig:HumanEvalResults}. There is unfortunately no statistically significant difference between the baseline Dual Encoder ($M=2.97$, $SD=1.27$) and the one combined with the weighting model ($M = 3.04$, $SD = 1.27$), as established by a Wilcoxon rank-sum test, $W(1838) = 410360$, $p = 0.23$. These inconclusive results are probably due to the very low agreement between the evaluation participants (Krippendorff's $\alpha$ for continuous variable $= 0.36$). The fact that the lower and upper bounds are only separated by 2 standard deviations confirms the difficulty for the raters to discriminate between responses. We hypothesise that the nature of the corpus, which is heavily dependent on an external context (the movie scenes), makes it particularly difficult to assess the consistency of the responses. 

\iffalse
%The results of this human evaluation are presented in Table \ref{tab:HumanEvalResults}. There is unfortunately no statistically significant difference between the baseline Dual Encoder and the one combined with the weighting model ($p = 0.23$ using Wilcoxon rank sum test). These inconclusive results are probably due to the very low agreement between the evaluation participants (Krippendorff's $\alpha$ for a continuous variable $= 0.36$). The fact that the lower and upper bounds are only separated by 2 standard deviations confirms the difficulty for the raters to discriminate between responses. We hypothesise that the nature of the corpus, which is heavily dependent on an external context (the movie scenes), makes it particularly difficult to assess the consistency of the responses. 

\begin{table}[htbp]
    \centering
    \begin{tabular}{lrrrr} 
        \hline
        Model name               &    $M$ &   $SD$ &   $N$ \\ 
        \hline
        Random                   & $2.08$ & $1.04$ & $960$ \\
        Dual Encoder             & $2.97$ & $1.27$ & $960$ \\
        Dual Encoder + weighting & $3.04$ & $1.27$ & $960$ \\
        Authored                 & $4.05$ & $1.09$ & $960$ \\ 
        \hline
    \end{tabular}
    \caption{Average human ratings of the responses generated by the four models tested. $M$ = Mean consistency. $SD$ = Standard deviation. $N$ = Total number of ratings. Consistency evaluated on a scale of 1 to 5.}
    \label{tab:HumanEvalResults}
\end{table}
\fi

Some examples of responses produced by the two Dual Encoder models illustrate the improvements brought by the weighting model. In \ref{ex:continuation}, the baseline Dual Encoder selected a turn continuation rather than a reply, while the second model avoids this pitfall. Both \ref{ex:continuation} and \ref{ex:namedent} also show that the dual encoder with instance weighting tends to select utterances with fewer named entities.

\ex.\label{ex:continuation}
    \textit{Context of conversation:} \\
    -- This is General Ripper speaking. \\
    -- Yes, sir. \\
    -- Do you recognize my voice?" \\ 
    $\Rightarrow$ \textit{Response of Dual Encoder:} \\
    -- This is General Nikolas Pherides, Commander of the Third Army. I'm Oliver Davis. \\
    $\Rightarrow$ \textit{Response of Dual Encoder + weighting:} \\
    -- Yes, sir. I'm Gideon.
    
\ex.\label{ex:namedent}
    \textit{Context of conversation:} \\
    -- Let me finish dinner before you eat it... Chop the peppers...\\
    -- Are you all right? \\ 
    $\Rightarrow$ \textit{Response of Dual Encoder:} \\
    -- No thanks, not hungry. Harry Dunne. \\
    $\Rightarrow$ \textit{Response of Dual Encoder + weighting:} \\
    -- Yes I'm fine. Everything is ok.

\iffalse
%Table \ref{table:contrastive} illustrates some examples of responses produced by the two Dual Encoder models. In the first example, the baseline Dual Encoder selected a turn continuation rather than a reply, while the second model avoids this pitfall. Both the first and the second examples also show that the dual encoder with instance weighting tends to select utterances with fewer named entities.

\begin{table}[htbp]
    \centering
    \begin{tabular}{p{70mm}}
    \textbf{Context of conversation:} \\
    \ \ -- This is General Ripper speaking. \\
    \ \ -- Yes, sir. \\
    \ \ -- Do you recognize my voice?" \\ 
    $\Rightarrow$ \textit{Response of Dual Encoder:} \\
    \ \ -- This is General Nikolas Pherides, Commander of the Third Army. I'm Oliver Davis. \\
    $\Rightarrow$ \textit{Response of Dual Encoder + weighting:} \\
    \ \ -- Yes, sir. I'm Gideon.  \vspace{1mm} \\ 
     \vspace{-1mm}
    \textbf{Context of conversation:} \\
    \ \ -- Let me finish dinner before you eat it... \\
    Chop the peppers...\\
    \ \ -- Are you all right? \\ 
    $\Rightarrow$ \textit{Response of Dual Encoder:} \\
    \ \ -- No thanks, not hungry. Harry Dunne. \\
    $\Rightarrow$ \textit{Response of Dual Encoder + weighting:} \\
    \ \ -- Yes I'm fine. Everything is ok. \\ 
    \end{tabular}
    \caption{Contrastive examples of responses produced by the Dual Encoder models.}
    \label{table:contrastive}
\end{table}
\fi

\section{Discussion}
\label{sec:discussion}

The limitations of neural conversational models trained on large, noisy dialogue corpora such as movie and TV subtitles have been discussed in several papers. Some of the issues raised in previous papers are the absence of turn segmentation in subtitling corpus \cite{vinyals2015neural,serban2015text,slt2016}, the lack of long-term consistency and ``personality'' in the generated responses \cite{li-EtAl:2016:P16-13}, and the ubiquity of dull, commonplace responses when training generative models \cite{li-EtAl:2016:N16-11}. To the best of our knowledge, this paper is the first to propose an instance weighting approach to address some of these limitations.  One related approach is described in \cite{DBLP:journals/corr/ZhangLWZ17} which also relies on domain adaptation for neural response generation, using a combination of online and offline human judgement. Their focus is, however, on the construction of personalised conversation models and not on instance weighting. 

The empirical results corroborate the hypothesis that assigning weights to the training examples of ``noisy'' dialogue corpora can boost the performance of neural conversation models.  In essence, the proposed approach replaces a one-pass training regime with a two-pass procedure: the first pass to determine the quality of each example pair, and a second pass to update the model based on the observed pair and its associated weight. We also showed that these weights can be determined in a data-driven manner with a neural model trained on example pairs selected for their adherence to specific quality criteria.

Instead of this two-pass procedure, an alternative approach is to directly learn a conversation model on the subset of example pairs that are known to be of high-quality. However, one major shortcoming of this approach is that it considerably limits the size of the training set that can be exploited. For instance, the data used to estimate the weighting model in Section \ref{sec:datasets} corresponds to a mere 0.1 \% of the total English-language part of the OpenSubtitles corpus (since the utterances had to be associated with speaker names derived from aligned scripts in order to apply the heuristics). In contract, the proposed two-pass procedure can scale to datasets of any size. 

The results from Section \ref{sec:evaluation} are limited to retrieval-based models. One important question for future work is to investigate whether the results carry over to generative, sequence-to-sequence models. As generative models are more computationally intensive to train than retrieval models, the presented approach may bring another important benefit, namely the ability to filter out part of the training data to concentrate the training time on ``interesting'' examples with a high cohesion between the context and its response.

%\note{ \cite{liu-EtAl:2016:EMNLP20163} have shown that current metrics for dialogue systems rarely correlate with human judgement (...) }

\section{Conclusion}
\label{sec:conclusion}

Dialogue corpora such as chat logs or movie subtitles are very useful resources for developing open-domain conversation models. They do, however, also raise a number of challenges for conversation modelling. Two notable challenges are the lack of segmentation in dialogue turns (at least for the movie subtitles) and the presence of external context that is not captured in the dialogue transcripts themselves (leading to mentions of person names and  unresolvable named entities).

This paper showed how to mitigate these challenges through the use of a \textit{weighting model} applied on the training examples. This weighting model can be estimated in a data-driven manner, by providing example of ``high-quality'' training pairs along with random pairs extracted from the same corpus. The criteria that determine how these training pairs should be selected depend in practice on the type of conversational model one wishes to learn. This instance weighting approach can be viewed as a form of \textit{domain adaptation}, where the data points from the source domain (in this case, the original corpus) are re-weighted to improve the model performance in a target domain (in this case, the interactions in which the conversation model will be deployed).

Evaluation results on retrieval-based neural models demonstrate the potential of this approach. The weighting model is essentially a preprocessing step and can therefore be combined with any type of conversational model. 

Future work will focus on two directions. The first is to extend the weighting model to account for other criteria, such as ensuring diversity of responses and coherence across turns. The second is to evaluate the approach on other types of neural conversational models, and more particularly on generative models. 

%\nocite{*}

\bibliographystyle{acl_natbib}
\bibliography{biblio2017}

\end{document}